\documentclass[11pt]{article}

\usepackage[preprint]{acl}
\usepackage[table]{xcolor}
\usepackage{times}
\usepackage{latexsym}
\usepackage[T1]{fontenc}

\usepackage[utf8]{inputenc}

\usepackage{microtype}

\usepackage{inconsolata}

\usepackage{subcaption}
\usepackage{graphicx}
\usepackage{xcolor}
\usepackage{geometry}
\usepackage{booktabs}
\usepackage{multicol}
\usepackage{multirow}
\usepackage[raster, most]{tcolorbox}
\usepackage{array}

\usepackage{makecell}

\title{Mary, the Cheeseburger-Eating Vegetarian:

Do LLMs Recognize Incoherence in Narratives?}

\author{
\parbox{\linewidth}{\center
Karin de Langis\textsuperscript{1}, Püren Öncel\textsuperscript{2}, Ryan Peters\textsuperscript{1}, Andrew Elfenbein\textsuperscript{4}, \\
Laura Kristen Allen\textsuperscript{2}, Andreas Schramm\textsuperscript{3}, \& Dongyeop Kang\textsuperscript{1} }\\
\textsuperscript{1}Department of Computer Science and Engineering, University of Minnesota \\
\textsuperscript{2}Department of Educational Psychology, University of Minnesota \\
\textsuperscript{3}Department of Linguistics, Hamline University \\
\textsuperscript{4}Department of English, University of Minnesota \\
\texttt{dento019@umn.edu}
}

\begin{document}
\maketitle
\begin{abstract}
Leveraging a dataset of paired narratives, we investigate the extent to which large language models (LLMs) can reliably separate incoherent and coherent stories.
A probing study finds that LLMs' internal representations can reliably identify incoherent narratives. 
However, LLMs generate responses to rating questions that fail to satisfactorily separate the coherent and incoherent narratives across several prompt variations, hinting at a gap in LLM's understanding of storytelling.  
The reasoning LLMs tested do not eliminate these deficits, indicating that thought strings may not be able to fully address the discrepancy between model internal state and behavior.
Additionally, we find that LLMs appear to be more sensitive to incoherence resulting from an event that violates the setting (e.g., a rainy day in the desert) than to incoherence arising from a character violating an established trait (e.g., Mary, a vegetarian, later orders a cheeseburger), suggesting that LLMs may rely more on prototypical world knowledge than building meaning-based narrative coherence.
The consistent asymmetry found in our results suggests that LLMs do not have a complete grasp on narrative coherence.

\end{abstract}

\section{Introduction}
Contemporary models of (human) reading comprehension characterize comprehension as a dynamic process in which the reader continually builds and updates representations to maintain coherence and integrate new information with prior knowledge, e.g., \cite{kintsch1998comprehension, myers1998accessing, van1999construction}. Specifically, narratives evoke mental simulations of events \cite{kintsch1998comprehension}, and when processing narratives, readers focus on dimensions such as causality, time, space, motivations and protagonist to maintain coherence \cite{magliano1999role, mcnamara2009toward, zwaan1998situation}. 

\begin{figure}[t]
    \centering
    \includegraphics[width=\linewidth]{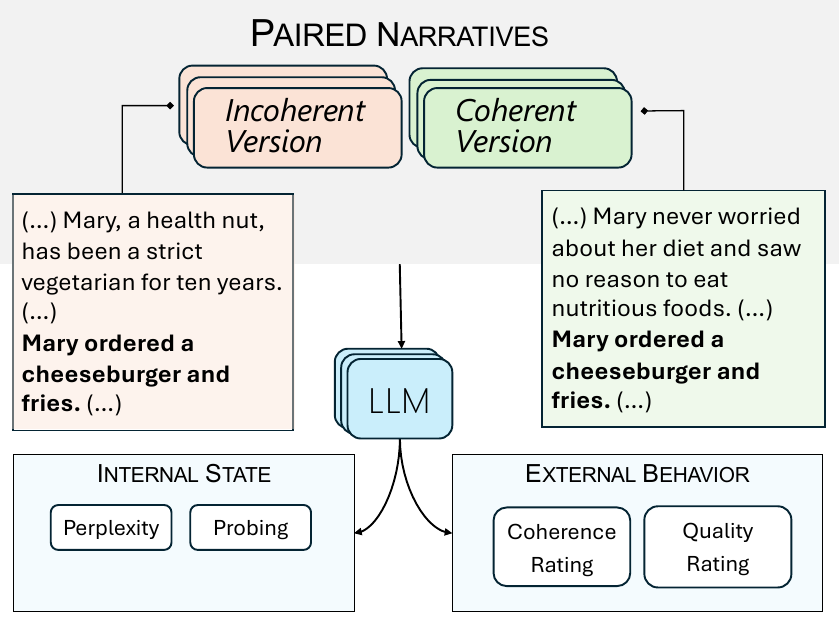}
    \caption{We use paired coherent and incoherent narratives to investigate the extent to which LLMs maintain coherence when tracking fictional entities. Internal measures differentiate coherent from incoherent stories at the incoherent location, yet LLMs' explicit coherent ratings at the end of the story fail to do so.
    }
    \label{fig:fig1}
\end{figure}

For LLMs, claims of ``narrative understanding'' rely on whether an analogue of this coherence-monitoring process is present. Standard evaluations rarely isolate incoherence itself, confounding it with topic, difficulty, or style; as a result, they cannot adjudicate whether models register contradictions or rely on surface heuristics. This raises the question: \textit{Are LLMs able to reliably recognize coherence in narratives?} To explore this question, we leverage a longstanding paradigm in human reading studies: paired narratives that are identical except for a critical inconsistency manipulation \citep{albrecht1993updating}. Specifically, we adapt and augment materials from \citet{rizzella2024prospective} to study LLMs, enabling precise comparison of LLM responses to coherent versus incoherent narratives (Figure~\ref{fig:fig1}). 

We first examine \textit{internal} LLM representations of the narratives using token-level perplexity and a probing study of hidden states.
Both measures indicate that LLMs can internally distinguish between the coherent and incoherent narratives, at least at the location where incoherence occurs (or does not occur).
However, LLMs appear \textbf{more sensitive to event-setting incoherence} (violations of commonsense expectations about the scene, like being given chopsticks and and then being served pizza) than to \textbf{trait-behavior incoherence} (violations of a character's established personality or trains, like a vegetarian ordering a cheeseburger).
Human readers show no such asymmetry, suggesting that LLMs may be relying more on background world knowledge to identify incoherence.




Secondly, we examine \textit{external} LLM responses by asking LLMs to assess the coherence and quality of each story. 
Surprisingly, both instruction-tuned and reasoning models fail to reliably distinguish coherent from incoherent narratives on both Likert ratings and forced choice true or false questions.
Examination of reasoning traces reveals that LLMs often apply muddled logic -- for instance, by identifying innocuous details as incoherent -- revealing \textbf{a discrepancy between their internal sensitivity to narrative structure and their explicit understanding of it}.
This dissociation between internal and external indicators of LLM narrative understanding highlights a potential vulnerability: models may internally register indicators of narrative incoherence without being able to reliably recognize or report it when instructed to do so.


As LLMs are increasingly deployed in real-world applications, and especially in educational settings focused on literature and creative writing, understanding how they construct or fail to construct narrative coherence is essential for evaluating their trustworthiness and alignment with human narrative reasoning. We believe that our findings, together with the paired narrative dataset introduced here, provide a valuable foundation for future research on the mechanisms of narrative comprehension in LLMs.

\section{Related Work}
Our work intersects with previous research in comprehension processes in both humans and language models. We cover each of these separately in this section.

\subsection{Comprehension processes in humans}
Human reading comprehension involves constructing mental representations of the discourse content \cite{gernsbacher1997attentuating,kintsch1998comprehension, myers1998accessing}. As reader encounter new information, they integrate it with their prior knowledge (both prior discourse and general world knowledge). Because of the limited capacity of the reader's working memory, comprehension proceeds in cycles in which only a small portion of the text is processed at a time \cite{kintsch1978toward,garrod1981bridging,van1983strategies}.  If this comprehension process is successful, the resulting representation forms a situation model, a network of semantic and causal relations that connects information from the discourse with reader's prior knowledge.

The construction of a \textit{coherent} situation model is the central goal of comprehension, achieved by integrating new information in both local and global contexts \cite{glenberg1992comprehension, o1992comprehension}. Readers establish local coherence by integrating incoming information with the immediately preceding context (i.e., information in their short-term memory), and global coherence by linking incoming information to earlier, no-longer-active but relevant information \cite{albrecht1993updating, glenberg1992comprehension, o1992comprehension}. Researchers have long debated whether inferences support only local or also global coherence in comprehension (e.g., \citet{mckoon1992inference}). \citet{albrecht1993updating} found that the reader detects global incoherence in narratives immediately even when local coherence is maintained, indicating that despite capacity limitations in working memory, the reader maintains access to global situation models throughout the course of a narrative and continually integrates new information into this situation model during reading. This can be partially but not completely mitigated by qualifying information (i.e., the inclusion of information that explains away the apparent incoherence). Further nuances have been explored in the literature, such as when past information that is no longer relevant conflicts \cite{obrien1998updating}, or when prospective information conflicts \cite{rizzella2024prospective}.

The present work uses the incoherent narratives studied by \citet{rizzella2024prospective}, which contributed to understanding how and when people integrate new information into mental representations of narratives as they read.

\subsection{Coherence and comprehension in LLMs}

Narrative comprehension in LLMs has been characterized as coherent, but brittle \citep{chang2024language}, and there are debates as to the extent that robust situation modeling is possible under transformer-only architectures \cite{mahowald2024dissociating}. One obvious challenge in situation modeling for LLMs is context window limitations: a model can only process a finite number of tokens, making it impossible to process long narratives such as those spanning multiple novels. While there are approaches that address context window limitations \cite{su2024roformer}, such approaches are computationally expensive. Efficient, human-like solutions to processing long narratives include hierarchical processing by incremental summarization (e.g., \citet{song2024hierarchical}). A second, more pernicious challenge is that LLMs have been found to make nonhuman-like mistakes in situation modeling, even when the context length poses no challenges. For example, generations can refer to non-existent discourse entities: ``I doubt Michael baked a cake. It's in the oven,'' \citep{schuster2022sentence}, and LLMs can produce highly incoherent narrative summaries \cite{goyal2022snac}. Early evidence also suggests that LLMs have inconsistent abilities in making pragmatic inferences in texts and inferring relations between events \cite{chang2024language, delangis2025llms}. 

Alongside studies analyzing LLM outputs to infer model knowledge and capabilities, a parallel branch of research answers questions about what LLMs truly understand by examining the information encoded in their internal representations. Probing classifiers are one such approach, revealing what knowledge is linearly accessible within networks \cite{alain-2018, belinkov-2022}.
One key question in this line of research is to what extent transformers trained on sequential prediction tasks develop coherent internal world models of the processes generating their input sequences. \citet{li2023emergent} showed that a GPT variant trained to predict legal moves in the Othello game developed an emergent, and linear \cite{nanda-etal-2023-emergent}, internal representation of the board state, despite never being explicitly given this information. 
Extending this to discourse, entity tracking studies find that maintaining entity states during narratives is fundamental to comprehension, yet some LLMs struggle with this \cite{kim-shuster-2023}. Notably, only models pretrained on both text and code demonstrated robust entity tracking, suggesting that text-only pretraining might be insufficient \cite{kim-shuster-2023}

\section{Paired Narrative Datasets}
Our experiments use a dataset of paired coherent and incoherent stories based on materials from a reading psychology study by \citet{rizzella2024prospective}. The dataset contains 18 paired stories originally designed by psychologists for narrative comprehension research and undergone normative testing in human participants. 

\begin{figure}[t]
    \centering
    \includegraphics[width=\linewidth]{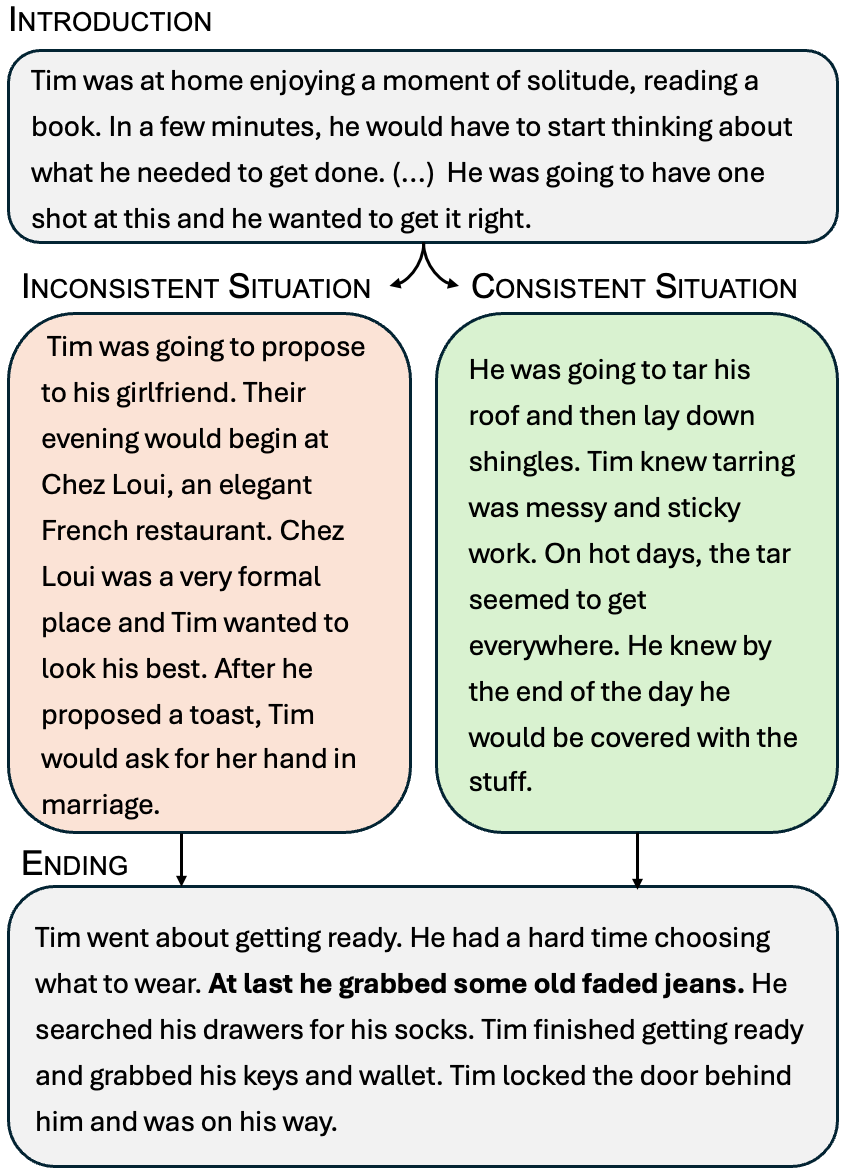}
    \caption{An exemplary incoherent-coherent story pair. Both versions share an identical introduction and conclusion, but differ in the middle situation: the incoherent version includes the inconsistent situation. Note the boldface \textbf{target} sentence, which conflicts with the inconsistent situation only. Some introductory text is omitted for brevity, denoted by (...).}
    \label{fig:narrative-structure}
\end{figure}

We chose this data set because it is the only dataset using the contradiction paradigm that has publicly available human psychometric data to our knowledge, thus enabling fine-grained comparison between human and LLM responses. Second, the incoherent scenarios presented in this dataset are well-studied; the majority were first conceived by \citet{albrecht1993updating} and have been replicated in multiple studies of human readers.

\subsection{Story structure}
All stories follow a fixed structure, as shown in Figure~\ref{fig:narrative-structure}. Each coherent-incoherent pair consists of two almost completely identical stories, with the exception of the situation introduced after the introduction. We denote the consistent and inconsistent situations as $s_c$ and $s_i$, respectively, and the critical target sentence as $t$. In the incoherent version, $t$ conflicts with $s_i$, but notice that in the coherent version, $t$ remains compatible with $s_c$. This design isolates the point of incoherence, which enables precise analysis of model and human responses to coherence violations. 


\subsection{Incoherence taxonomy and annotation}
Incoherence occurs when it becomes impossible to hold one consistent interpretation of the text, but there are many possible sources of this interpretive breakdown \cite{reinhart1980conditions}. In particular, we note two distinct patterns in the dataset:
\begin{itemize}
    \item \textit{Trait-behavior conflict:} occurs when a strongly established character trait is later violated by that character's behavior. For example, in Figure~\ref{fig:fig1}, Mary's established vegetarian character trait conflicts with her later choice to order cheeseburger. 
    \item \textit{Event-setting conflict:} occurs when an event contradicts a strict expectation established by the narrative's setting or social context. 
    For example, in Figure~\ref{fig:narrative-structure}, social expectations about the established setting of a formal restaurant conflicts with Tim dressing in old faded jeans.
\end{itemize}

\begin{figure}[t]
    \centering
    \includegraphics[width=\linewidth]{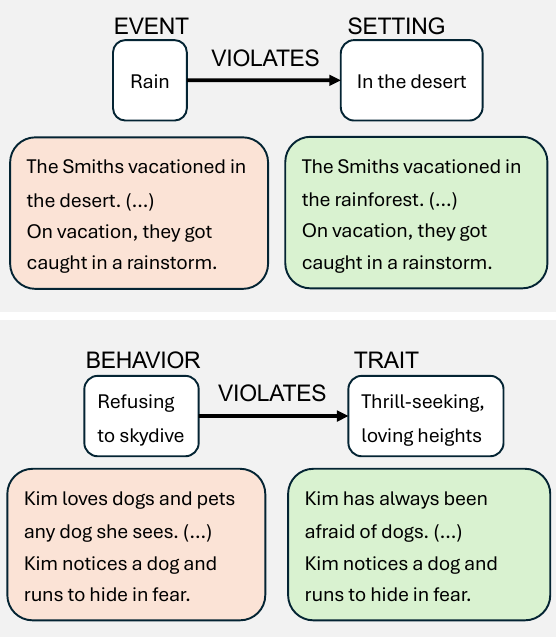}
    \caption{We identify and annotate two types of incoherence in the dataset: event-setting and trait-behavior. They differ in the target of the violation; specifically, trait-behavior is incoherent because it violates a strongly established character trait, while event-setting is incoherent because it violates some facet of the established narrative scene (e.g., the geographical location or the social norms).}
    \label{fig:conflict}
\end{figure}

Figure \ref{fig:conflict} also illustrates these two types of incoherence. 
We focus on this distinction because we hypothesize that LLMs may approach these two types of conflicts differently.
Event-setting conflicts center on violation of the narrative situational context.
Because all stories are set in realistic environments, such conflicts may engage the model's pretrained world knowledge.
In contrast, trait-behavior conflicts are centered around a character whose personality traits are defined only within the story itself, which may limit the influence of background knowledge and instead rely on internal narrative consistency.

To obtain ground truth labels for the incoherence type, we have three collaborators independently label each story's conflict type based on the above definitions. The inter-annotator agreement is calculated with Krippendorff's $\alpha = 0.835$, indicating strong agreement. Among the incoherent stories, 40\% were even-setting conflicts ($n = 10$), and 60\% were trait-behavior ($n = 15$). 

\subsection{Dataset augmentation}\label{sec:story-augments}
The original materials from \citet{rizzella2024prospective} include 18 story pairs. 
To improve the reliability and precision of our conclusions, we augment it with seven additional pairs, resulting in a larger sample size of 25 story pairs (50 stories total).

The new stories are generated using a human-in-the-loop writing process with Claude \cite{anthropic_claude_2025}, followed by iterative refinement incorporating feedback from psychologists. 
All experimental results shown, unless otherwise stated, use the augmented dataset. The reported findings hold for both the original and augmented datasets, and experimental results for the original stories alone can be found in Appendix.

\begin{figure*}
    \centering
    \includegraphics[width=\linewidth]{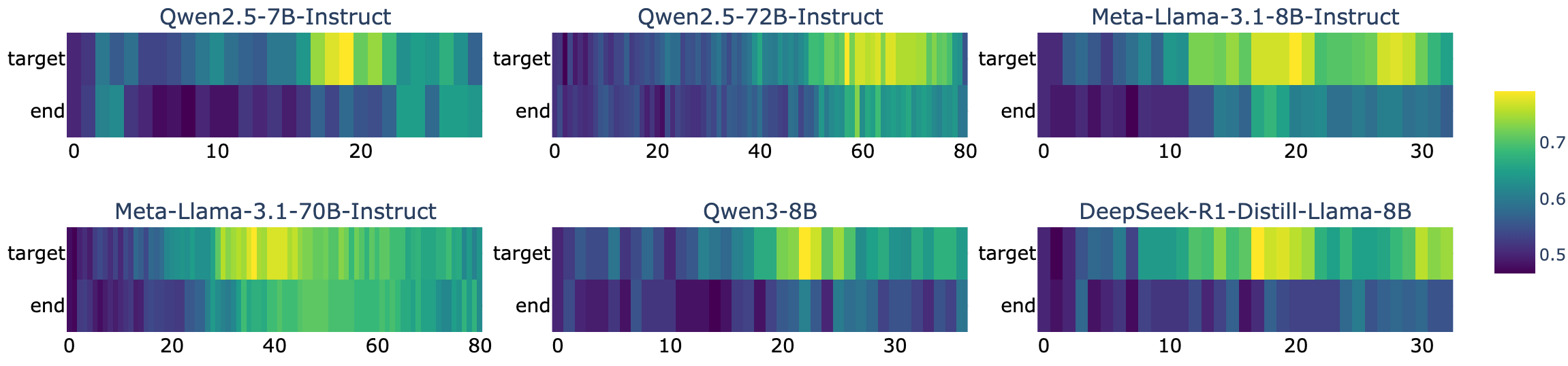}
    \caption{Mean accuracies across 10-fold cross-validation for probing hidden state representations to identify incoherent narratives. We probe both at the end of the target sentence that contains the incoherence in the incoherent story version (target), and at the conclusion of the story (end). The $x$ axis denotes model layer. All models show strong separation at the target location, but by the story's end, separation is notably weaker, with smaller models in particular near chance ($\approx 50 - 60\%$ accuracy). Llama3.1-70B has the best performance at the story's end, and it also demonstrates the best understanding of coherence in responses to rating questions.}
    \label{fig:probe}
\end{figure*}

\begin{figure}[t]
    \centering
        \includegraphics[width=\linewidth,trim=0 0.3cm 0 0, clip]{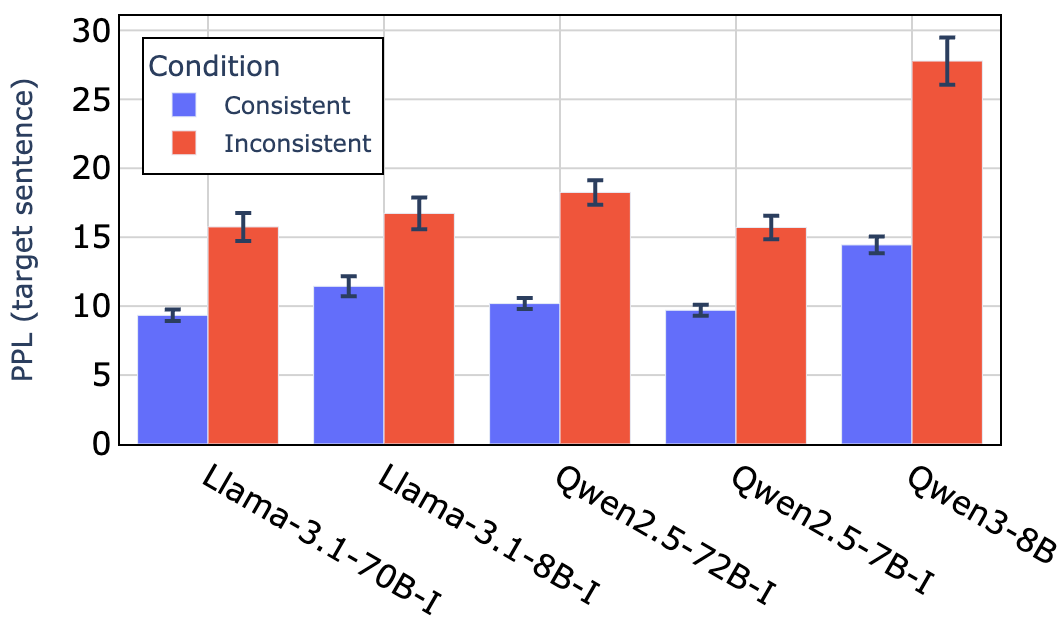}
    \vspace{-8mm}
    \caption{LLMs assign higher perplexity (i.e., lower likelihood) to incoherent events relative to coherent ones.}
    \label{fig:perplexity}
\end{figure}

\section{Experimental Setup and Methods}
All experiments follow a consistent procedure for model inference and, in prompting experiments, systematic variation of prompts.

\paragraph{Model Inference.}
We use four open-source instruction-tuned LLMs across two model families and sizes: Llama3.1 with 8B and 70B parameters \citep{dubey2024llama}, and Qwen2.5 with 7B and 72B parameters \citep{yang2024qwen2}. We also test two reasoning variants from each model family: a Llama-8B model distilled from DeepSeek-R1 \citep{deepseekai2025deepseekr1} and Qwen3-8B with reasoning enabled. We use the transformers library \citep{wolf-etal-2020-transformers} for model inference, applying 4-bit quantization \cite{dettmers2023case} to the larger models to meet computational constraints. Otherwise, default model configurations and generation hyperparameters are used in all cases. All experiments are completed on a Linux server equipped with one Nvidia A40 GPU (48GB). 

\paragraph{Prompt Variation.}
Our experiments involving model generation systematically vary prompts and instructions to capture a more robust range of model responses. Specifically, we use \textsc{FormatSpread} to vary punctuation and whitespace \cite{sclar2024quantifying}, and we write three paraphrases of each instruction. The combination of these prompt variations yields 30 prompts per stimulus, giving us a better estimate of LLM performance. Specific prompts used can be found in \S\ref{sec:prompts}.


\section{Results}
To answer our central question \textit{Are LLMs able to reliably recognize incoherence in narratives?}, we perform extensive analyses:
We first investigate how models encode narrative coherence internally (\S\ref{sec:ppl}, \S\ref{sec:probe}) and then examine their explicit behavioral judgments (\S\ref{sec:coherence_quality}, \S\ref{sec:reasoning}). Finally, we compare model sensitivity across the two incoherence types and relate these findings to human data (\S\ref{sec:conflict-types_llm-vs-human}).

\begin{table*}[t]
\centering
\small
\begin{subtable}[t]{0.48\textwidth}
\centering
\begin{tabular}{lccc}
\toprule
Model & \shortstack{Con. Rating\\(max 7)} & \shortstack{Incon. Rating\\(max 7)} & Diff. ($\uparrow$)\\
\midrule
\textit{Instruction-tuned} \\
$\rightarrow$ Llama3.1 8B  & 6.26 (0.10) & 6.03 (0.11) & 0.00 \\
$\rightarrow$ Llama3.1 70B & 6.47 (0.08) & 4.61 (0.16) & \textbf{1.66} \\
$\rightarrow$ Qwen2.5 7B   & 6.63 (0.06) & 6.61 (0.06) & 0.04  \\
$\rightarrow$ Qwen2.5 72B  & 6.21 (0.08) & 5.53 (0.11) & 0.73 \\
\midrule
\textit{Reasoning-enabled} \\
\textbf{*} Qwen3 8B   & 6.17 (0.11) & 5.00 (0.14) & \underline{1.18} \\
\textbf{*} R1-Distill-Llama 8B  & 6.17 (0.06) & 5.97 (0.07) & 0.28\\
\bottomrule
\end{tabular}
\caption{Mean LLM coherence ratings.}
\label{tab:coherence_ratings}
\end{subtable}
\hfill
\begin{subtable}[t]{0.48\textwidth}
\centering
\begin{tabular}{ccc}
\toprule
\shortstack{Con. Rating\\(max 7)} & \shortstack{Incon. Rating\\(max 7)} & Diff. ($\uparrow$) \\
\midrule
\\
4.49 (0.09) & 4.39 (0.11) & 0.10 \\
4.97 (0.06) & 4.58 (0.07) & \textbf{0.39} \\
5.35 (0.12) & 5.16 (0.11) & 0.19 \\
4.71 (0.06) & 4.50 (0.06) & 0.21 \\
\midrule
\\
4.06 (0.16) & 3.75 (0.17) & \underline{0.31} \\
5.15 (0.06) & 5.09 (0.07) & 0.06 \\
\bottomrule
\end{tabular}
\caption{Mean LLM quality ratings (aligned rows).}
\label{tab:quality_ratings}
\end{subtable}

\caption{Comparison of model ratings for (a) narrative coherence and (b) narrative quality across coherent and incoherent story versions. 
Ratings use a 7-point Likert scale (7 = fully coherent / highest quality, 4 = neutral). 
Each value reflects the mean across 30 prompt trials per story. Reasoning models are marked with \textbf{*}.}
\label{tab:combined_ratings}
\end{table*}


\subsection{LLMs assign lower likelihoods to incoherent events}\label{sec:ppl}
We begin by testing whether LLMs recognize incoherent events as less probable within the narrative. For each story pair, we compute the perplexity of the target sentence $t$ given the story context: $\text{PPL}(t | s_i)$ and $\text{PPL}(t | s_c)$ for the incoherent and coherent versions, respectively.
Across all models, $\text{PPL}(t | s_i) > \text{PPL}(t | s_c)$, indicating that LLMs systematically estimate incoherent continuations as less likely or higher perplexity (Figure~\ref{fig:perplexity}).
This result provides initial evidence that LLMs encode a sensitivity to narrative coherence at the token-probability level.

\subsection{Incoherence is recoverable from target hidden state}\label{sec:probe}
To test whether this coherence sensitivity is explicitly represented within model activations, we conduct a probing study \cite{belinkov-2022, alain-2018}. We frame detection of incoherence as a binary classification task and fit a linear probe on representations extracted from the hidden state of each layer at the last token in $t$.
Probes are evaluated with 10-fold cross-validation at the \textit{paired} story level, ensuring that coherent and incoherent versions of a story always appear in the same fold. 

Probes at the last token of the target sentence achieve high accuracy (~80\%) across models, confirming that coherence information is linearly decodable from internal representations (Figure~\ref{fig:probe}). The most informative layers tend to be mid-to-upper layers, consistent with prior findings on semantic integration \cite{hewitt2019structural}.
When probing the final story token, accuracy tends to dramatically decline, especially for smaller models. It is possible that coherence distinctions decay as the model processes later narrative content. Alternatively, coherence information may be encoded locally in the residual stream at the target token position and but remain accessible via attention mechanisms. The information would then not automatically be propagated forward to subsequent tokens, leading to decreased probing performance when probing the last sentence.

\subsection{Explicit ratings of coherence and quality show weak behavioral sensitivity}\label{sec:coherence_quality}

\paragraph{Coherence Rating.} We next assess whether the models' explicit judgments reflect their internal sensitivity. Each model rates story coherence on a 7-point Likert scale (7 = fully coherent), and we sample ratings across 30 perturbed prompts per story.
Contrary to their internal representations, most models show minimal behavioral distinction between the two story types: the best-performing model (Llama3.1-70B) achieves a mean 1.66 point separation between coherent and incoherent stories, and four out of six models have less than one point mean difference between coherent and incoherent stories (Table~\ref{tab:combined_ratings} (a)).  
Thus, while models encode coherence internally, this sensitivity is at best weakly expressed in their overt ratings.

\begin{figure}
    \centering
    \includegraphics[width=\linewidth]{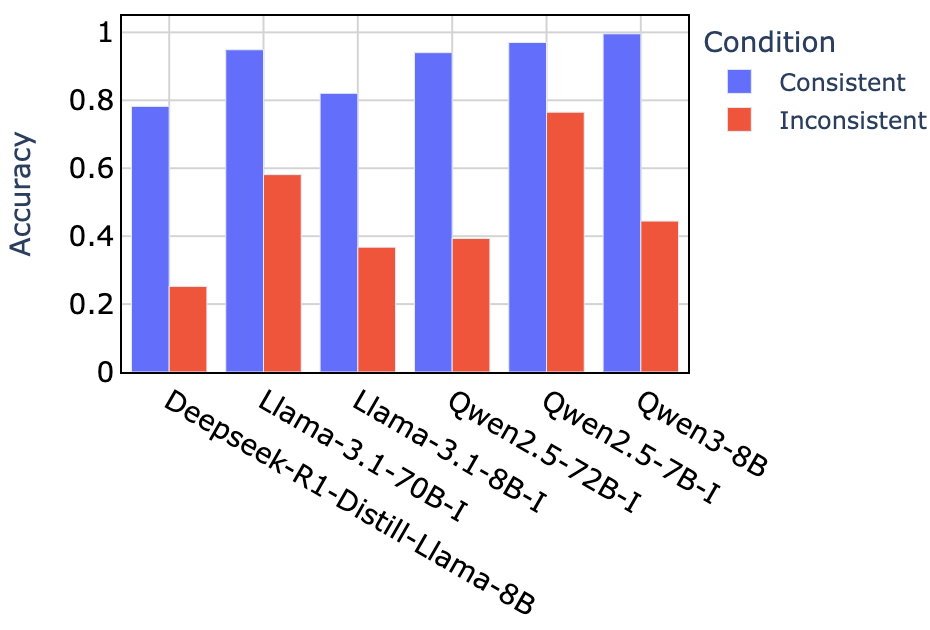}
    \caption{Model accuracies by condition when responding True or False to whether a story is coherent.}
    \label{fig:coherencetf}
\end{figure}

Since it is possible that in a forced-choice scenario LLMs will show a stronger behavioral distinction between incoherent and coherent narratives, we additionally prompt models with a true or false evaluation of story coherence, but we do not find that this improves model performance (Figure ~\ref{fig:coherencetf}).

\paragraph{Quality Rating.} When prompted to rate narrative quality, models again exhibit limited differentiation: no model shows more than a one-point mean difference between coherent and incoherent versions (Table~\ref{tab:combined_ratings}b).
We do find that quality and coherence ratings have a weak to moderate positive correlation across all models (Pearson's $r = 0.33, p < 0.05$), indicating that models do associate the two concepts, as expected.


\begin{table*}[htbp]
\centering
\small
\setlength{\tabcolsep}{3pt}
\begin{tabular}{llccccc}
\toprule
 & & 
\makecell{\textbf{Probe}\\\textbf{Target}} & 
\makecell{\textbf{Probe}\\\textbf{End}} & 
\makecell{\textbf{Perplexity}\\\textbf{Target}} & 
\makecell{\textbf{Coherence}\\\textbf{Response}} & 
\makecell{\textbf{Quality}\\\textbf{Rating}} \\
\midrule
\multirow{2}{*}{\textbf{Llama3.1}} 
& 8B-Instruct & \underline{0.850} & 0.667 & \textbf{0.880} & 0.594 & 0.595 \\
& 70B-Instruct & \textbf{0.942} & 0.817 & \underline{0.880} & 0.765 & 0.672 \\
\midrule
\multirow{2}{*}{\textbf{Qwen2.5}}
& 7B-Instruct & \textbf{0.800} & 0.683 & \underline{0.689} & 0.720 & 0.561 \\
& 72B-Instruct & \textbf{0.958} & 0.817 & \underline{0.830} & 0.707 & 0.621 \\
\midrule
\multirow{2}{*}{\textbf{Reasoning}} 
& R1-Distill-Llama-8B & \textbf{0.850} & 0.600 & \underline{0.845} & 0.517 & 0.470 \\
& Qwen3-8B & \textbf{0.842} & 0.625 & \underline{0.772} & 0.737 & 0.414 \\
\bottomrule
\end{tabular}
\caption{Accuracy of various approaches in separating incoherent and coherent stories with LLMs. Measures based on internal model representations—perplexity and probe—work best, but these rely on \textit{a priori} knowledge of the incoherent target sentence. Coherence ratings are only above chance for large models.}
\label{tab:model_metrics}
\end{table*}

\subsection{Reasoning models reveal partial metacognitive awareness}\label{sec:reasoning}
LLMs may be lacking in the meta-cognition required to transform an internal detection of incoherence into a coherence or quality rating.
Reasoning-enabled variants are designed to generated explicit ``thought strings'' before answering.
Thus, we test whether this reflective reasoning leads to stronger behavioral distinctions between the consistent and inconsistent stories.

While reasoning models occasionally verbalize the link between quality and coherence, their final ratings do not substantially improve (c.f. Qwen3-8B and Qwen2.5-7B, or Llama3.1-8B and R1-Distill-Llama-8B in Table~\ref{tab:combined_ratings}).
\begin{tcolorbox}[colback=gray!5, colframe=gray!50, 
                  boxrule=0.5pt, arc=2mm, left=4pt, right=4pt, top=3pt, bottom=3pt,title=Qwen3-8B thought (emphasis added)]
\noindent So, the question is about the quality. What makes a story high quality? Usually, it's about \textbf{coherence}, character development, plot, and whether it conveys a clear message or has a purpose. Let me break it down.
\end{tcolorbox}


Moreover, these models sometimes produce false positives when attempting to detect incoherence, such as misinterpreting consistent details as contradictions or hallucinating nonexistent incoherent events. This suggests that although reasoning prompts activate coherence-related concepts, they do not reliably translate internal sensitivity into accurate explicit judgments., e.g.:
\begin{tcolorbox}[colback=gray!5, colframe=gray!50, 
                  boxrule=0.5pt, arc=2mm, left=4pt, right=4pt, top=3pt, bottom=3pt,title=Qwen3-8B thought (false positive )]
\noindent I need to check if there's any inconsistency. The story mentions Bill does additional workouts before and after his walks. That's a bit odd because if he's walking, maybe he's doing it as part of his exercise. 
\end{tcolorbox}

\subsection{Comparison to Human Readers: Sensitivity to Incoherence Types}\label{sec:conflict-types_llm-vs-human}
\begin{figure*}[t]
    \centering
    \includegraphics[width=\linewidth]{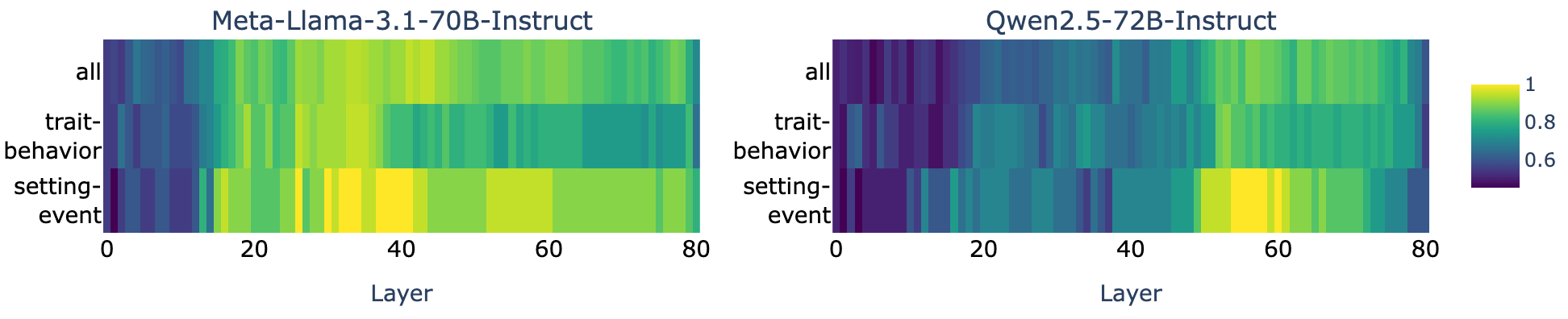}
    \vspace{-8mm}
    \caption{Probe accuracy in detecting incoherence using a linear regression train on hidden states, averaged across 10-fold cross validation. We conduct a probing study over three datasets: (1) only trait-behavior incoherence stories, (2) only setting-event incoherence stories. Probes generally achieve better separation for the setting-event stories.
    }
    \label{fig:probe-types}
\end{figure*}

Across analyses, models show a tendency toward \textit{greater sensitivity to event–setting incoherence} compared to trait-behavior incoherence.
Perplexity differences are larger for event–setting conflicts (Figure ~\ref{fig:ppl-type}), and probe accuracies are also consistently higher for this type (Figure~\ref{fig:probe-types}).
This pattern aligns with our hypothesis that event–setting incoherence leads to higher engagement of pretrained world knowledge, whereas trait–behavior incoherence depends more on within-story reasoning, which models handle less consistently.

In contrast, we do not find evidence that human readers show a similar sensitivity.
We examine human reading times\footnote{There is a well-established relationship between language model surprisal and human reading times \cite{wilcox2020predictive, oh2022comparison}.} of the target sentence of each story using the publicly available data collected by \citet{rizzella2024prospective}. 
On average, reading times were similar for setting-event stories ($M=1967.91, SD=478.80$) and trait-behavior stories ($M=1913.62, SD=487.55$). Reading times are measured in milliseconds. A paired-samples t-test confirmed this difference was not statistically significant, $t(29) = 0.65, p = .523, M_{\text{diff}}= 54.29, 95\% CI = [-117.39, 225.97]$. Since the human study was not designed to test story type effects, individual participants did not necessarily read all possible combinations of story type and consistency conditions (e.g., inconsistent \texttt{x} setting-event). Therefore, we do not test for possible interaction effects between story category and consistency due to limited statistical power.

\section{Discussion}

Claims that LLMs truly understand narratives ultimately hinge on whether they can monitor and maintain coherence across a story. Our experiments aim to shed light on whether LLMs actually recognize contradictions within a narrative or simply respond to shallow statistical patterns.

Our findings indicate that, even for the relatively short and simple narratives studied here, the LLMs tested -- especially the smaller ones at 7 - 8B parameters -- have difficulty reliably separating incoherent and coherent narratives.
Additionally, all LLMs tested fail to consistently identify coherent stories as higher quality, raising questions about LLM reliability on tasks involving understanding or evaluating storytelling. Despite these poor responses on various rating questions, probing does indicate that models' internal representations distinguish between coherent and incoherent stories, bringing \textit{why} they fail to give reasonable coherence and quality ratings into question.

One possible contributing factor is suggested by the drop in performance between probes on the token at the end of the target incoherence location and probes on the token ending the story: while LLMs may temporarily represent incoherence, they may loose access to that representation at later locations in the story. Interestingly, Llama-70B has the highest performance on the story-end probe and also has the best rating responses, which supports this interpretation.

Another possible explanation for the discrepancy between model internal states and responses is that the model representations are in fact capturing an attribute highly associated with incoherence, such as unexpectedness. Because our investigations of model internal state rely on \textit{a priori} knowledge of the incoherent target sentence and, in the case of probing, supervised learning, such a confounding factor could influence these results more strongly than the behavioral results.
While coherent events are unexpected, not all unexpected events are necessarily incoherent; in fact, unexpected events are oftentimes the key to a good story \cite{kintsch1980learning}. 
Thus, LLMs confounding unexpectedness and incoherence would be highly undesirable.
While our dataset does not allow us to investigate this, as a proper study would require carefully designed paired unexpected-incoherent stories, we believe that disentangling unexpectedness and incoherence in LLMs is an important direction for future work. 

Additionally, we note an apparent tendency for models to respond more to incoherence based on event-setting conflicts, rather than trait-behavior conflicts.
This is puzzling as humans do not appear to share this tendency.\footnote{The result hints at low understanding of character incoherence; it may align with an earlier finding that language model summaries make the most coherence errors with respect to character references \cite{goyal2022snac}.} 
It is possible that background world knowledge modulates the models' response heavily in some event-setting instances. 
Future work investigating, e.g., how LLMs respond to event-setting conflicts in fantasy worlds with narrative-specific rules would be another important direction for isolating this phenomenon.






\section{Conclusion}

We investigate how well six LLMs identify incoherence in narratives using a combination of metrics based on model internal representations and responses to prompts. 
To facilitate our experiments, we adapt and augment a paired narrative dataset originally used in human research.
Our results indicate that while internal representations achieve high -- albeit imperfect -- separation of incoherent and coherent narratives at the location of incoherence, this separation largely disappears by the end of the story. We also find that LLM responses identify the incoherence at generally lower rates than we find via perplexity or probing at the target. Together, our results suggest that LLMs lack a robust coherence-monitoring process during narrative comprehension.

\section{Limitations}
Our work studies narratives made for psychological studies rather than narratives drawn from naturally occurring stories. This also necessarily means that the dataset is limited in size, as it is hand-crafted. The advantage of this approach is that the stories are well designed to isolate the attribute of interest (coherence), but the results must be interpreted with the above caveats in mind.

Our analysis also does not investigate any proprietary models, as we rely on access to model hidden states. In a similar vein, while we examine six models, we cannot be certain our results generalize more broadly.

\section{Ethics Statement}
Our analyses of human daya use only publicly available data that do not contain personally identifiable information.

Additionally, the application of LLMs in creative domains such as literature raises ethical concerns regarding the potential displacement of human labor and creative expression. There is a risk of societal impact if LLMs begin to assume roles traditionally held by human writers and storytellers.

\bibliography{custom}

\appendix

\section{Appendix}

\subsection{Prompting Experiment Supplemental Materials}

\subsection{Prompts}\label{sec:prompts}
We use \textsc{FormatSpread} to vary whitespace and punctuation of stimuli data within the prompt elements in the following 10 ways:

\begin{itemize}

\item 
Story: (|STORY|)\\
(|QUESTION|)

\item 
Field: \{\}\\
Answer: \{\}\\
Story: (|STORY|)\\
(|QUESTION|)

\item 
Field: \{\} <sep>
Story: (|STORY|) <sep> (|QUESTION|)

\item 
Field - \{\}.
Story - (|STORY|). (|QUESTION|)

\item 
Field\t\{\}. 
Story\t(|STORY|). (|QUESTION|)

\item 
FIELD- \{\} \\
STORY- (|STORY|)\\
(|QUESTION|)

\item 
field:: \{\} \\
story:: (|STORY|) -- (|QUESTION|)

\item 
field - \{\} \\
story - (|STORY|) ,  (|QUESTION|)

\item 
Field\\
\t\{\}\\
\\
\t\{\}\\
Story\\
\t(|STORY|)\\
\t(|QUESTION|)

\item 
Field - \{\}\\
Story - (|STORY|)\\
(|QUESTION|)

\end{itemize}

\subsection{Coherence Rating Prompts}
Our three paraphrasings for coherence prompts (used in conjunction with above format spread to create 30 prompts).
\begin{itemize}
    \item In this activity, you will be presented with a story to read. Afterwards, you'll be asked a question about the story.
    
    QUESTION: How coherent or consistent does the story seem on a scale from 1-7, with 1 being very incoherent and 7 being very coherent.
    \item In this exercise, you will go through a narrative, and upon finishing, you'll answer a question about how good the story way.

    QUESTION: Rate the quality of the story on a scale from 1-7, with 1 being poor coherence and 7 being high coherence. Respond only with a number between 1 and 7.
    \item In this task you will read a story. At the end you will be asked a question about the story.

    QUESTION: How coherent or consistent does the story seem on a scale from 1-7, with 1 being very incoherent and 7 being very coherent.
\end{itemize}
Quality prompts follow this format, replace ``coherent'' with quality and ``incoherent'' with low quality.

\subsection{Coherence Forced Choice Prompts}
Our three paraphrasings for coherence forced choice prompts (used in conjunction with above format spread to create 30 prompts).
\begin{itemize}
    \item In this activity, you will be presented with a story to read. Afterwards, you'll be asked a question about the story. 
    
    Story: (|STORY|)
    
    Question: True or False: the story seems coherent and consistent. Only answer with True or False.
    \item In this exercise, you will go through a narrative, and upon finishing, you'll answer a question about how good the story way.
    \item In this task you will read a story. At the end you will be asked a question about the story.
\end{itemize}



\subsection{Perplexity differences}
\begin{figure}
    \centering
    \includegraphics[width=\linewidth]{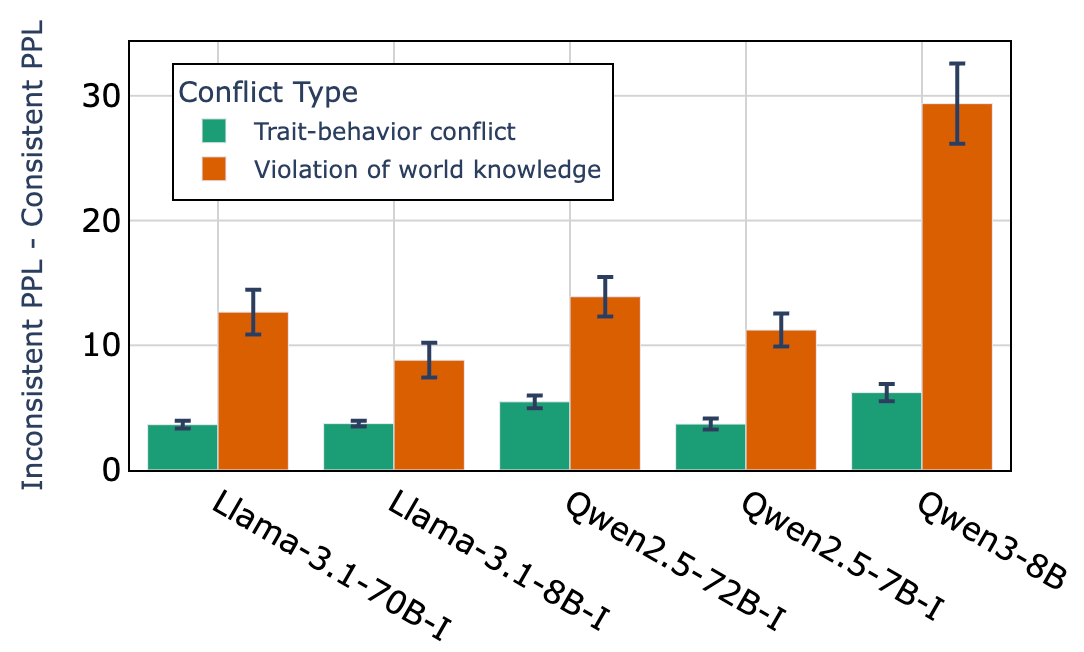}
    \caption{We find that there is a higher difference in perplexity for stories with event-setting incoherence.}
    \label{fig:ppl-type}
\end{figure}

We find that perplexity differences are higher for stories with event-setting incoherence (Fig.~\ref{fig:ppl-type}).

\section{No Augmentations}
Our main results with no augmentations to the dataset. Coherence ratings Table~\ref{tab:supp-coherence-ratings}, quality ratings \ref{tab:supp-qual-rating}.
Perplexity info is in Table~\ref{tab:conflict_ppl}

\begin{table}[h!]
\centering
\small
\begin{tabular}{lcc}
\hline
\textbf{Model} & \textbf{Coherent} & \textbf{Incoherent} \\
\hline
Distill-Llama-8B & $6.142 \pm 0.073$ & $5.937 \pm 0.082$ \\
Llama-3.1-70B-I               & $6.434 \pm 0.094$ & $4.681 \pm 0.176$ \\
Llama-3.1-8B-I                & $6.240 \pm 0.111$ & $6.006 \pm 0.131$ \\
Qwen2.5-72B-I                 & $6.150 \pm 0.091$ & $5.624 \pm 0.126$ \\
Qwen2.5-7B-I                  & $6.649 \pm 0.061$ & $6.628 \pm 0.064$ \\
Qwen3-8B                      & $6.091 \pm 0.127$ & $4.947 \pm 0.153$ \\
\hline
\end{tabular}
\caption{Mean coherence ratings on original stories only.}
\label{tab:supp-coherence-ratings}
\end{table}

\begin{table}[h!]
\centering
\small
\begin{tabular}{lcc}
\hline
\textbf{Model} & \textbf{Coherent} & \textbf{Incoherent} \\
\hline
Gemma-2-27Bi        & $4.037 \pm 0.091$ & $3.852 \pm 0.109$ \\
Llama-3.1-70B-I     & $4.685 \pm 0.065$ & $4.426 \pm 0.080$ \\
Llama-3.1-8B-I      & $4.875 \pm 0.119$ & $4.456 \pm 0.107$ \\
Qwen2.5-72B-I       & $4.562 \pm 0.062$ & $4.352 \pm 0.071$ \\
Qwen2.5-7B-I        & $5.150 \pm 0.060$ & $5.006 \pm 0.071$ \\
Qwen3-8B            & $4.056 \pm 0.164$ & $3.747 \pm 0.166$ \\
\hline
\end{tabular}
\caption{Mean quality ratings on original stories only.}
\label{tab:supp-qual-rating}
\end{table}

\begin{table}[h!]
\centering
\small
\begin{tabular}{lccc}
\hline
\textbf{Model} & \textbf{Conflict Type} & \textbf{Coherent PPL} & \textbf{Incoherent PPL} \\
\hline
Distill-Llama-8B & Trait-behavior conflict        & $12.113 \pm 1.159$ & $22.253 \pm 2.062$ \\
Distill-Llama-8B & Violation of world knowledge   & $39.761 \pm 7.062$ & $84.328 \pm 21.749$ \\
Llama-3.1-70B-I              & Trait-behavior conflict        & $6.251 \pm 0.456$ & $11.650 \pm 1.128$ \\
Llama-3.1-70B-I              & Violation of world knowledge   & $15.958 \pm 2.360$ & $31.253 \pm 6.545$ \\
Llama-3.1-8B-I               & Trait-behavior conflict        & $6.472 \pm 0.492$ & $11.272 \pm 0.749$ \\
Llama-3.1-8B-I               & Violation of world knowledge   & $22.053 \pm 4.347$ & $35.819 \pm 7.532$ \\
Qwen2.5-72B-I                & Trait-behavior conflict        & $6.101 \pm 0.357$ & $13.286 \pm 1.071$ \\
Qwen2.5-72B-I                & Violation of world knowledge   & $14.509 \pm 1.615$ & $27.609 \pm 4.989$ \\
Qwen2.5-7B-I                 & Trait-behavior conflict        & $6.224 \pm 0.368$ & $9.297 \pm 0.790$ \\
Qwen2.5-7B-I                 & Violation of world knowledge   & $16.371 \pm 1.831$ & $25.058 \pm 4.517$ \\
Qwen3-8B                     & Trait-behavior conflict        & $9.733 \pm 0.882$ & $16.768 \pm 1.789$ \\
Qwen3-8B                     & Violation of world knowledge   & $23.747 \pm 2.223$ & $49.794 \pm 8.290$ \\
\hline
\end{tabular}
\caption{Non-augmented dataset perplexity means across coherent and incoherent stories for different types of incoherence.}
\label{tab:conflict_ppl}
\end{table}

\end{document}